\documentclass[preprint,12pt]{elsarticle}
\usepackage[utf8]{inputenc}
\usepackage{textcomp}

\usepackage{booktabs}      
\usepackage{multirow}      
\usepackage{amssymb}
\usepackage{makecell} 
\usepackage{rotating}

\usepackage{graphicx}   

\usepackage[dvipsnames]{xcolor}

\usepackage{amssymb}
\usepackage{amsmath}

\journal{Medical Image Analysis}

\begin{document}

\begin{frontmatter}



\title{Rank-Aware Agglomeration of Foundation Models for Immunohistochemistry Image Cell Counting}




\author[label1,label2]{Zuqi Huang\fnref{fn1}}
\author[label3]{Mengxin Tian\fnref{fn1}}
\author[label4]{Huan Liu\corref{cor1}}
\author[label1]{Wentao Li}
\author[label5]{Baobao Liang}
\author[label6]{Jie Wu}
\author[label7]{Fang Yan}
\author[label3,label8]{Zhaoqing Tang\corref{cor1}}
\author[label2]{Zhongyu Li\corref{cor1}}
\affiliation[label1]{
    organization={School of Software Engineering, Xi’an Jiaotong University},
    city={Xi’an},
    postcode={710049}, 
    state={Shaanxi},
    country={China}
}

\affiliation[label2]{
    organization={School of Computer Science, Shanghai Jiao Tong University},
    city={Shanghai},
    postcode={200240}, 
    country={China}
}
            
\affiliation[label3]{
    organization={Department of Gastrointestinal Surgery and the Gastric Cancer Center, Zhongshan Hospital, Fudan University},
    city={Shanghai},
    postcode={200032}, 
    country={China}
}

\affiliation[label4]{
    organization={School of Computer Science and Technology, Xi’an Jiaotong University},
    city={Xi’an},
    postcode={710049}, 
    state={Shaanxi},
    country={China}
}
\affiliation[label5]{
    organization={The Comprehensive Breast  Care Center, The Second Affiliated Hospital of Xi'an Jiaotong University},
    city={Xi’an},
    postcode={710114}, 
    state={Shaanxi},
    country={China}
}
\affiliation[label6]{
    organization={Department of Pathology, The Second Affiliated Hospital of Xi’an Jiaotong University},
    city={Xi’an},
    postcode={710114}, 
    state={Shaanxi},
    country={China}
}
\affiliation[label7]{
    organization={Shanghai Artificial Intelligence Laboratory},
    city={Shanghai},
    country={China}
}
\affiliation[label8]{
    organization={Department of General Surgery, Zhongshan Hospital (Xiamen), Fudan University},
    city={Xiamen},
    postcode={361015}, 
    state={Fujian},
    country={China}
}

\fntext[fn1]{Equal contribution.}
\cortext[cor1]{Corresponding authors: Zhongyu Li (zhongyuli@sjtu.edu.cn), Zhaoqing Tang (tang.zhaoqing@zs-hospital.sh.cn), Huan Liu (huanliu@xjtu.edu.cn).}

\begin{abstract}
Accurate cell counting in immunohistochemistry (IHC) images is critical for quantifying protein expression and aiding cancer diagnosis. However, the task remains challenging due to the chromogen overlap, variable biomarker staining, and diverse cellular morphologies. Regression-based counting methods offer advantages over detection-based ones in handling overlapped cells, yet rarely support end-to-end multi-class counting. Moreover, the potential of foundation models remains largely underexplored in this paradigm. To address these limitations, we propose a rank-aware agglomeration framework that selectively distills knowledge from multiple strong foundation models, leveraging their complementary representations to handle IHC heterogeneity and obtain a compact yet effective student model, CountIHC. Unlike prior task-agnostic agglomeration strategies that either treat all teachers equally or rely on feature similarity, we design a Rank-Aware Teacher Selecting (RATS) strategy that models global-to-local patch rankings to assess each teacher’s inherent counting capacity and enable sample-wise teacher selection. For multi-class cell counting, we introduce a fine-tuning stage that reformulates the task as vision–language alignment. Discrete semantic anchors derived from structured text prompts encode both category and quantity information, guiding the regression of class-specific density maps and improving counting for overlapping cells. Extensive experiments demonstrate that CountIHC surpasses state-of-the-art methods across 12 IHC biomarkers and 5 tissue types, while exhibiting high agreement with pathologists' assessments. Its effectiveness on H\&E-stained data further confirms the scalability of the proposed method.
\end{abstract}




\end{frontmatter}


\section{Introduction}
Immunohistochemistry (IHC) enables the visualization of disease-specific protein expression, such as Ki67, PD-L1, and CD3, and plays a vital role in cancer subtyping \cite{inamura2018update07}, targeted therapy selection \cite{song2019understanding02,zhang2022genetic03,casulo2019durvalumab04}, and prognosis assessment \cite{xu2018immune05,qiu2019prognostic06}. Quantitative interpretation of IHC results is critical for clinical decision-making. For example, the tumor proportion score (TPS) \cite{boyer2021pembrolizumab08}, which estimates the fraction of tumor cells expressing a specific biomarker, is widely used to guide treatment for patients above defined thresholds \cite{brattoli2024universal09}. This requires accurate enumeration of relevant cells to ensure reproducible and objective analysis. Manual counting by pathologists is tedious, time-consuming, and subject to inter-observer variability. The inherent complexity of IHC images, including diverse biomarker staining, high chromogen overlaps, and variations in cellular morphology, density, and spatial expression across tissue regions, further complicates this task. These challenges call for automated cell counting methods that are both accurate and robust under real-world IHC conditions.

Existing cell counting methods can be broadly categorized into detection-based \cite{ghahremani2022deep01, horst2024cellvit26, stringer2025cellpose3, archit2025segment28, paulauskaite2019deep30, he2017mask31, graham2019hover32} and regression-based methods \cite{he2021deeply13, xie2018microscopy25, zheng2024rethinking33}. Detection-based methods identify individual cell instances to infer counts but often degrade in densely packed regions and rely on costly instance annotations. In contrast, regression-based methods estimate spatial density maps and integrate them to obtain counts, enabling training with lower-cost point annotations and showing superior performance in overlapping-cell scenario \cite{he2021deeply13}. Despite these strengths, IHC scoring typically requires separate enumeration of biomarker-positive and -negative tumor cells rather than a single aggregated count, while most regression-based models rarely support such end-to-end multi-class counting and the use of foundation models (FMs) within this paradigm remains largely underexplored.

In computational pathology, a growing number of studies \cite{wang2022transformer14, huang2023visual15, zimmermann2024virchow216, hoptimus117, chen2024towards18} have focused on pretraining foundation models to capture the inherent representations of histopathological images. Due to heterogeneity in their architectures, training objectives, and data sources, different FMs often exhibit complementary strengths on specific data \cite{ma2024towards19}. To harness the complementary knowledge of multiple FMs, several works \cite{ma2024towards19, ranzinger2024radio20, sariyildiz2024unic21, sariyildiz2025dune23} have explored agglomeration strategies that distill the knowledge of multiple FMs into a single student model. Yet, existing approaches typically lack task-awareness or interpretability. They often treat all teachers equally or select them based solely on feature similarity to the student, without explicitly evaluating their task-specific effectiveness. As a result, they fail to provide fine-grained, performance-driven teacher selection during distillation, and this aspect has not yet been investigated in cell counting tasks.

To address the aforementioned limitations, we propose a rank-aware agglomeration framework that distills knowledge from multiple FMs into a compact student, CountIHC, capable of retaining or surpassing the counting performance of the strongest teacher FM, while substantially reducing model complexity. Through agglomeration, CountIHC selectively inherits the complementary and generalizable representations of FMs to handle heterogeneous staining and morphological patterns in IHC. Specifically, a Rank-Aware Teacher Selecting (RATS) strategy is introduced, leveraging an unsupervised global-to-local patch ranking task modeling ordinal relationships in cell quantity. By evaluating teachers on their inherent ability to preserve the correct ranking within each patch group, RATS dynamically selects the optimal teacher. To achieve end-to-end multi-class counting, we design a fine-tuning stage that reformulates counting as vision–language alignment, where discrete semantic anchors derived from structured text prompts encode both category and quantity information. Aligning these anchors with image features regresses class-specific density maps, ensuring accurate estimation even in cell overlapping regions. Our contributions are summarized as follows:
\begin{itemize}
    \item To the best of our knowledge, it is the first work to apply multiple foundation models agglomeration to IHC cell counting. We propose a rank-aware agglomeration framework that produces a compact model, CountIHC, which retains or surpasses the best-performing teacher FM while substantially reducing model complexity.
    \item For foundation model agglomeration, we design Rank-Aware Teacher Selecting (RATS), a task-driven teacher selection strategy that employs an unsupervised global-to-local patch ranking task. Aligned with the cell counting objective, it assesses the inherent counting capability of each teacher and enables fine-grained distillation from the optimal one.
    \item A fine-tuning stage is introduced for multi-class cell counting, where discrete semantic anchors are constructed through structured text prompts to encode both cell category and quantity information, and are aligned with image features to regress class-specific density maps, enhancing inter-class distinction and multi-class counting accuracy.
    \item Extensive experiments on multi-center IHC datasets with 12 biomarker and 5 tissue types demonstrate the effectiveness of CountIHC, which outperforms state-of-the-art methods and shows strong agreement with pathologists’ assessments. The method also exhibits scalability when applied to H\&E-stained data.
\end{itemize}

\section{Related work}

\subsection{Cell counting}
Accurate quantification of cells in microscopic images is critically important in both medical and biological domains \cite{guo2019sau24}. In recent years, inspired by advances in counting algorithms from computer vision, deep learning-based cell counting has gained considerable momentum, aiming to alleviate the tedious burden of manual annotation for domain experts. Existing methods can be broadly grouped into two types: detection-based \cite{ghahremani2022deep01, horst2024cellvit26, stringer2025cellpose3, archit2025segment28, paulauskaite2019deep30, he2017mask31, graham2019hover32} and regression-based methods \cite{he2021deeply13, xie2018microscopy25, zheng2024rethinking33}. Detection-based methods estimate cell counts by identifying individual instances using annotations such as bounding boxes or segmentation masks. Paulauskaite-Taraseviciene et al. \cite{paulauskaite2019deep30}, for example, explored the use of Mask R-CNN \cite{he2017mask31} for overlapping nuclei detection through a two-stage procedure involving potential region proposal and joint mask prediction. Recent methods often incorporate cell-specific proxy maps, such as the vector flow in Cellpose \cite{stringer2025cellpose3} and the distance map in HoVerNet \cite{graham2019hover32}. CellViT \cite{horst2024cellvit26} further leverages large-scale pretrained models to achieve high-precision instance segmentation of cell nuclei in whole-slide images (WSIs). However, these detection-based methods still face challenges in counting cells that are densely clustered, strongly adhesive, and embedded within complex histopathological structures \cite{he2021deeply13}.

Regression-based cell counting methods have attracted increasing attention due to their superior ability to handle overlapping cells and their reliance on low-cost point-level annotations. These methods predict spatial cell density maps and infer the total count by integrating or summing over the entire predicted map. Additionally, the local maxima in the density map can serve as approximations of cell centroids. He et al. \cite{he2021deeply13} extended the FCRN-based density regression framework \cite{xie2018microscopy25} by incorporating deep supervision to facilitate intermediate feature learning. Zheng et al. \cite{zheng2024rethinking33} proposed a decoupled architecture that separates counting and localization into two dedicated branches and introduces global context modeling, achieving sate-of-the-art performance across multiple cell counting benchmarks. Recently, studies \cite{ma2024clip34, jiang2023clip35, liang2023crowdclip36} have explored the use of vision-language models such as CLIP \cite{radford2021learning37} for crowd counting, using text prompts to guide density estimation. Although promising in natural scene applications, their applicability to cell or nuclei counting remains underexplored. Moreover, most existing methods do not support simultaneous multi-class cell counting within a single image, limiting their applicability in scenarios where distinct cell categories must be quantified independently.

IHC scoring requires accurate counting of relevant cells or nuclei to assess protein expression, thereby supporting clinical decision-making \cite{ghahremani2022deep01}. Recent efforts have increasingly focused on automating biomarker evaluation using cell-level detection, segmentation, and scoring techniques. For Ki-67, BCData \cite{huang2020bcdata47} offers a large-scale benchmark dataset that enables proliferation index estimation based on positive and negative tumor cell annotations. DeepLIIF \cite{ghahremani2022deep01} leverages multiplex immunofluorescence as a pixel-level reference to jointly learn cell segmentation and expression prediction. Yan et al. \cite{yan2024artificial48} developed a PD-L1 scoring pipeline by integrating established detection and segmentation modules with rule-based metrics to compute clinical indices (e.g., TPS). Brattoli et al. \cite{brattoli2024universal09} proposed a universal analyzer for HER2, ER, and PR assessment, employing existing detection models to facilitate cross-marker generalization. While these approaches show promise in specific clinical settings, they are primarily detection-based and thus face challenges in managing densely packed or chromogen-overlapping cells, which are common in IHC images. Moreover, their applicability remains confined to limited biomarker and tissue settings, with insufficient validation on broader and more heterogeneous IHC datasets.

\subsection{Multiple foundation models agglomeration}
A wide range of foundation models (FMs) have recently emerged in computer vision and related fields. Trained on large-scale datasets, they exhibit strong generalization and broad applicability to diverse downstream tasks. For example, CLIP \cite{radford2021learning37}, trained on web-scale image–text pairs, achieves impressive zero-shot performance across visual benchmarks. In computational pathology, self-supervised pretraining methods such as DINO \cite{caron2021emerging38} and iBOT \cite{zhou2021ibot39} have been adopted to build pathology-specific FMs that reach state-of-the-art performance on downstream tasks \cite{zimmermann2024virchow216, hoptimus117, chen2024towards18, azizi2023robust40, vorontsov2024foundation41, xu2024whole42}. One representative example is Virchow2 \cite{zimmermann2024virchow216}, trained on 3.1 million whole-slide images (WSIs) covering both H\&E and IHC stains, which achieves strong results across multiple pathology benchmarks through domain-adapted self-supervised learning.

Although foundation models exhibit broad applicability, training them from scratch remains computationally expensive and time-consuming. Even when adapted to downstream tasks, large-scale variants are often required to sustain performance, creating heavy resource demands that hinder clinical deployment. To mitigate these challenges, knowledge distillation (KD) \cite{hinton2015distilling43} transfers the representational capacity of a large teacher to a compact student trained on the same domain data. It has been applied to computational pathology foundation models \cite{zimmermann2024virchow216, filiot2025distilling44}, although performance degradation often accompanies size reduction. This gap largely stems from the limited efficacy of conventional feature or label matching strategies \cite{zeng2025agglomerating45}.

Given the diversity of training data, learning strategies, and downstream performance across foundation models, recent studies \cite{ma2024towards19, ranzinger2024radio20, sariyildiz2024unic21, heinrich2025radiov222, sariyildiz2025dune23, zeng2025agglomerating45} have explored agglomerative distillation to unify multiple FMs and harness their complementary strengths for improved generalization. While aggregation, ensemble, and fusion approaches \cite{bilal2023aggregation_sup1, hou2016patch_sup2, luo2025ensemble, solorzano2024ensemble_sup4, chen2020pathomic_sup5} combine intermediate features or decision outputs, agglomeration distills the knowledge of multiple pretrained teachers into a single model, enabling a unified representation space derived from their learned competencies. AM-RADIO \cite{ranzinger2024radio20} first integrated representations from CLIP \cite{radford2021learning37}, DINO \cite{caron2021emerging38}, and SAM \cite{kirillov2023segment46}, achieving robust performance across heterogeneous vision tasks. RADIOv2.5 \cite{heinrich2025radiov222} further addressed resolution imbalance and token compression to enhance scalability in high-resolution domains. UNIC \cite{sariyildiz2024unic21} incorporated ladder projectors and stochastic teacher dropping, resulting in a universal encoder that achieves competitive results across multiple classification benchmarks. Extending this paradigm to computational pathology, GPFM \cite{ma2024towards19} employed unified knowledge distillation from multiple expert encoders, demonstrating improved generalizability across a wide spectrum of clinical applications.

Current agglomeration strategies predominantly fall into two categories: equal learning \cite{ma2024towards19, ranzinger2024radio20, heinrich2025radiov222} and teacher dropping regularization (tdrop) \cite{sariyildiz2024unic21, sariyildiz2025dune23}. In the former, the contributions of all teacher models are treated equally by averaging their respective distill losses. However, Sarıyıldız et al. \cite{sariyildiz2024unic21} observed that teachers do not contribute equally without further intervention. To mitigate this, they introduced a regularization strategy based on feature-level similarity: with the highest loss magnitude is retained during backpropagation, while the others are randomly discarded according to a predefined probability. Although student models derived from these strategies outperform teachers on specific tasks, the absence of explicit teacher capacity evaluation limits their ability to adaptively allocate supervision according to task-specific performance. They operate in a task-agnostic and opaque manner, ignoring model heterogeneity on a per-sample basis. 
\section{Methods}
\label{sec1}

\subsection{Overview}
\begin{figure}[ht] 
  \centering                   

  \includegraphics[
    width=0.95\linewidth,       
  ]{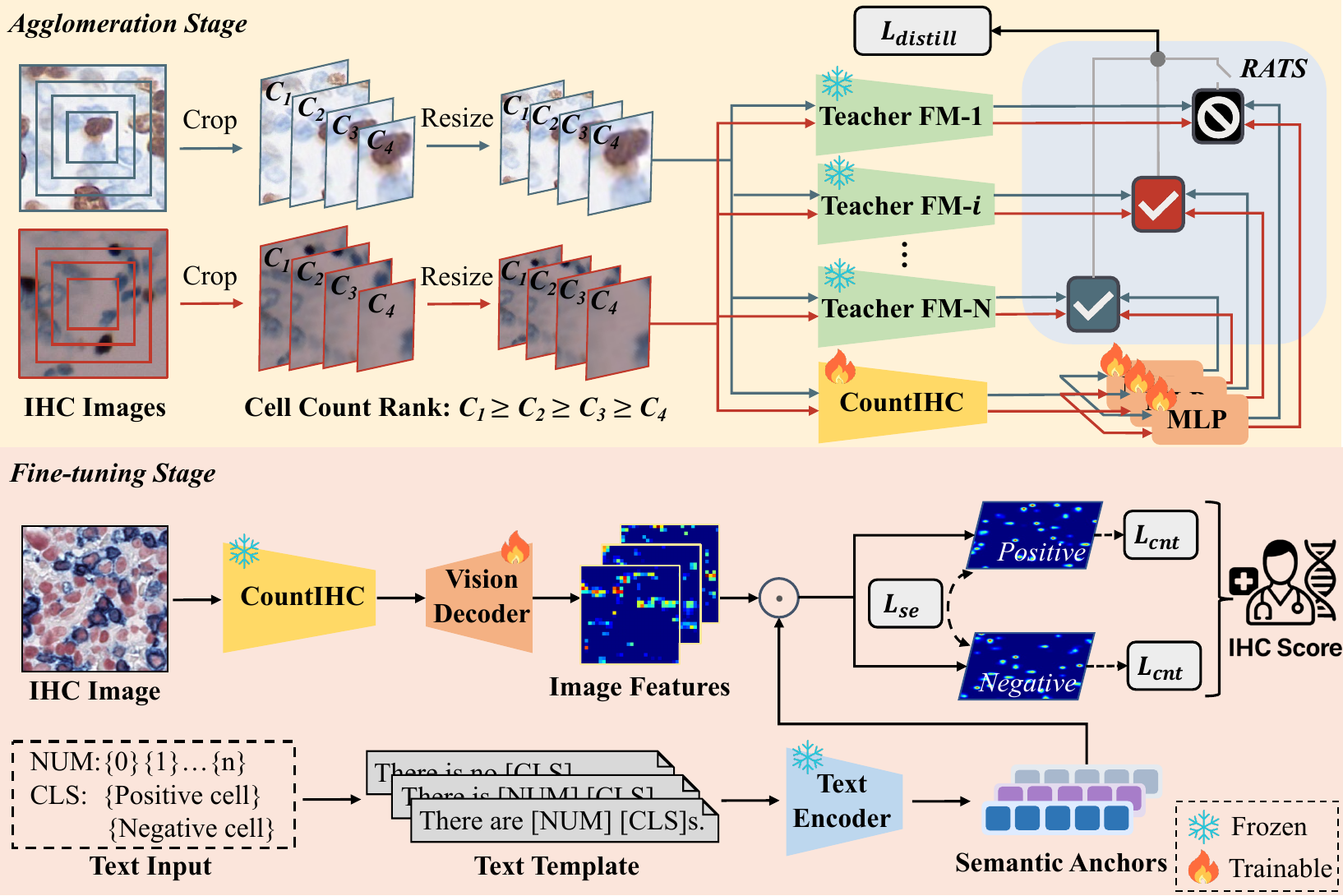}

  \caption{Overview of the proposed rank-aware agglomeration framework and anchor-guided fine-tuning for multi-class cell counting. }\label{fig:main}
\end{figure}
The overview of our proposed method is shown in Fig. \ref{fig:main}. It comprises two successive stages: a rank-aware agglomeration stage for task-driven and fine-grained learning from multiple foundation models, and an anchor-guided fine-tuning stage tailored for multi-class cell counting, with the incorporation of the text modality.

During the agglomeration stage, we propose a novel strategy termed Rank-Aware Teacher Selecting (RATS) to identify the optimal teacher foundation model and guide subsequent learning. Specifically, RATS evaluates multiple frozen foundation models on a task-relevant proxy: a global-to-local patch ranking task constructed from spatially ranked cropped regions of each input image. Without requiring annotations, this ordinal task provides a meaningful measure of each model’s inherent counting ability. RATS selects the optimal teacher sample-wise based on ranking consistency, and routes its knowledge into a compact student model via a distillation loss $L_{distill}$.

In the fine-tuning stage, we introduce discrete semantic anchors that encode both cell quantity (“NUM”) and category (“CLS”) information through structured text prompts. The alignment between visual features and these anchors enables anchor-guided density map generation across multiple categories. To further enhance the spatial separability of predicted density maps, we propose a Spatial Exclusivity Loss $L_{se}$, which penalizes inter-class co-activation and strengthens class-wise discriminability. The counts of positive and negative tumor cells derived from the predicted density maps are subsequently utilized for computing IHC scores, such as TPS or Ki-67 labeling index (LI).
\subsection{Agglomeration framework: setup and preliminaries}
Foundation models trained with different paradigms often exhibit complementary capabilities. For example, CLIP-style vision–language models emphasize semantic abstraction, while DINO-style self-distillation models better preserve spatial and structural fidelity \cite{radford2021learning37, oquab2023dinov2}. The goal of the agglomeration framework is to distill knowledge from a pool of $N$ teacher foundation models $\{ T_1, T_2, \ldots, T_N \}$ into a single student model $S$. In this study, we select two representative foundation models from computational pathology, Virchow2 \cite{zimmermann2024virchow216} and H-optimus-1 \cite{hoptimus117}, which are among the few trained with extensive IHC-stained WSIs and have demonstrated remarkable performance across various downstream benchmarks. The motivation is to aggregate their complementary representations and inherit their domain-specific capacity to model inherent IHC characteristics. In addition, we incorporate CLIP \cite{radford2021learning37} as a third teacher, inspired by its strong performance in recent crowd counting tasks \cite{ma2024clip34, jiang2023clip35, liang2023crowdclip36}. Our aim is to extend the paradigm of reformulating counting into a vision-language alignment problem to the IHC multi-class setting, which requires the student model to acquire multimodal alignment capabilities. To balance performance and computational efficiency for practical deployment, we adopt a compact ViT-Base \cite{dosovitskiy2020image49} architecture as the student encoder.

To enable feature-level alignment between the student model $S$ and each teacher $T_i$, we introduce a lightweight teacher-specific projector between $S$ and each $T_i$. Each projector is implemented as a two-layer multilayer perceptron (MLP) to adapt feature dimensions and facilitate knowledge transfer. Moreover, following findings from prior work \cite{sariyildiz2024unic21, heinrich2025radiov222}, we integrate intermediate representations from multiple layers, which have been shown to benefit downstream performance. Specifically, we apply a set of projectors to the outputs of selected intermediate layers and aggregate them to obtain the final output from the student. The alignment of the student and each teacher output, $S(x)$ and $T(x)$, is jointly supervised by cosine similarity and $\text{smooth-}\ell_1$ loss, as defined in \cite{ranzinger2024radio20}, formulated as:
\begin{equation}
\mathcal{L}_{\text{distill}} = \mathcal{L}_{\text{cos}}(S(x), T(x)) + \mathcal{L}_{\text{smooth-}\ell_1}(S(x), T(x))
\end{equation}
Regarding the agglomeration strategy, a straightforward way is to treat all teacher models equally by computing $\mathcal{L}_{\text{distill}}$ between the student and each teacher, then summing the losses across all teachers \cite{ma2024towards19, ranzinger2024radio20, heinrich2025radiov222}. However, as noted in \cite{sariyildiz2024unic21}, teachers do not contribute equally without further intervention. To address this, some approaches adopt teacher dropping based on feature-level similarity \cite{sariyildiz2024unic21, sariyildiz2025dune23}: they retain the teacher with the highest distillation loss and randomly discard others, aiming to balance the influence among multiple teachers. Despite this, such selection remains heuristic and lacks interpretability. It offers no principled mechanism for evaluating heterogeneous teachers based on their effectiveness on the target task and data. As a result, it fails to achieve fine-grained and task-specific guidance. This motivates us to design a more informed teacher selection strategy, introduced in the following subsection.
\subsection{Rank-Aware Teacher Selecting}
\begin{figure}[ht] 
  \centering                   

  \includegraphics[
    width=0.85\linewidth,       
  ]{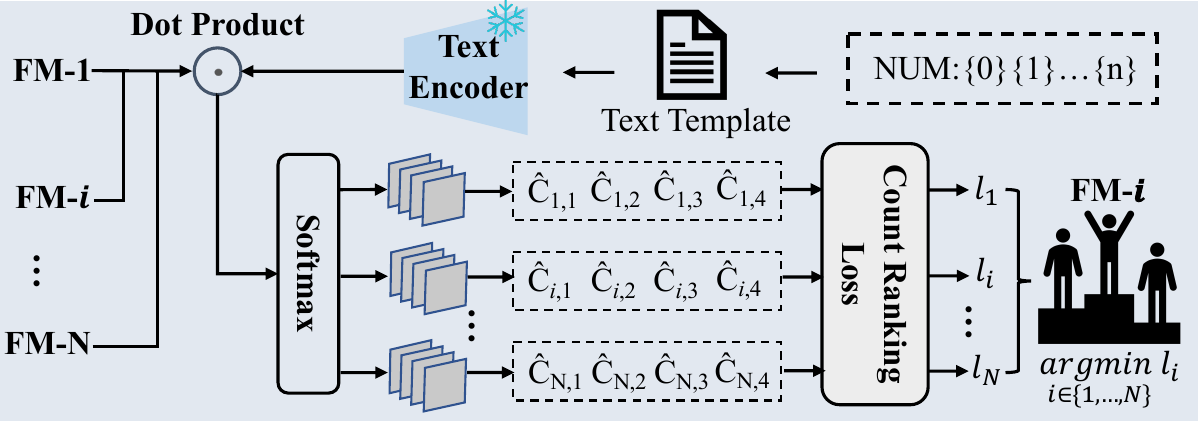}

  \caption{Illustration of the Rank-Aware Teacher Selecting (RATS) strategy. In an unsupervised setting, given a global-to-local cropped patch group, the output of each candidate foundation model (FM-$i$) is aligned with discrete count candidates to yield a predicted count $\hat{C}_{i,j}$ for patch $p_j$. A count-ranking loss serves as the rank-aware criterion, and the model that performs best by this criterion is selected as the group-specific teacher.}\label{fig:RATS}
\end{figure}
To provide a unified and label-free measure of each foundation model’s counting capacity on the same input, we introduce an unsupervised proxy task: global-to-local patch ranking. This auxiliary task is highly relevant to the underlying goal of cell counting, and enables teacher selection based on each model’s inherent ability to preserve ordinal relationships in spatially cropped patch groups.

To construct ordinal supervision without annotation, we leverage the prior that, under center-aligned cropping, larger patches tend to contain equal or greater numbers of cells compared to their nested sub-regions. Based on this observation, we generate ranked patch groups from input IHC images using a global-to-local cropping strategy. Formally, for each image, a sequence of $k$ patches $\{p_1, p_2, \ldots, p_k\}$ is constructed as follows:
\begin{enumerate}
    \item The largest patch $p_k$ is cropped from raw image using a predefined maximum size $M$.
    \item Centered at the geometric center of $p_1$, a sequence of progressively smaller patches $\{p_1, \ldots, p_{k-1}\}$ is extracted using a set of scale ratios $\{s_1, \ldots, s_{k-1}\} \subset (0,1),$ where each patch $p_i$ has size $M \times s_i$. This preserves the aspect ratio and avoids cell deformation or distortion.
    \item All cropped patches $\{p_i\}_{i=1}^{k}$ are then resized to a uniform resolution $M \times M$ for model input compatibility.
\end{enumerate}

To estimate the cell quantity in each patch, we sequentially feed the ranked patch group into each of the frozen teacher models. For a given teacher $T_i$, this produces a sequence of feature embeddings $\{y_{i,1}, y_{i,2}, \ldots, y_{i,k}\}$, where $y_{i,j} = T_i(p_j)$ represents the feature vector extracted from patch $p_j$. As illustrated in Fig. \ref{fig:RATS}, each embedding is compared against a predefined set of semantic anchors using cosine similarity, and a softmax operation is applied to obtain a probability distribution over discrete count candidates. The predicted count $\hat{C}_{i,j}$ for patch $p_j$ under teacher $T_i$ is then computed as the expectation of this distribution. The construction of semantic anchors and count prediction process will be detailed in subsection \ref{subsec1}. While the absolute values of predicted counts may be noisy under this unsupervised setting, our focus lies in modeling their ordinal relationship, emphasizing relative ordering rather than exact quantities.

We employ a count ranking loss to evaluate each teacher model’s inherent capacity for perceiving ordinal relationships among patches, thereby serving as a unified and interpretable criterion for teacher selection. Given a batch of $B$ patches divided into $G = B / k$ groups, where $g$-th group consists of a ranked patch list $\{p_1^{(g)}, p_2^{(g)}, \ldots, p_k^{(g)}\}$, the count ranking loss is formulated as:
\begin{equation}
\mathcal{L}_{\text{rank}} = \sum_{g=1}^{G} \sum_{(i,j) \in \mathcal{P}_g} \max\left(0, -(\hat{C}_j^{(g)} - \hat{C}_i^{(g)}) + \varepsilon \right),
\end{equation}
where $\hat{C}_i^{(g)}$ and $\hat{C}_j^{(g)}$ are the predicted cell counts for patches $p_i^{(g)}$ and $p_j^{(g)}$ within group $g$, and $\mathcal{P}_g = \{ (i, j) \mid 1 \leq i < j \leq k \}$ denotes the set of all ranked patch index pairs within group $g$. The ranking margin \( \varepsilon \) is set to zero. This loss penalizes cases where the predicted ranking contradicts the reference order, assuming without loss of generality that \( \hat{C}_j \geq \hat{C}_i \). The order follows the predefined global-to-local ranking prior, where patches with larger indices contain equal or larger numbers of cells.

For each teacher $T_i$, the ranking loss $\mathcal{L}_{\text{rank}}^i$ is computed as above, and the teacher selection follows the rank-aware criterion:
\begin{equation}
i^\star = \operatorname*{arg\,min}_{i \in \{1,\ldots,N\}} \mathcal{L}_{\text{rank}}^i,
\qquad
T_{\text{selected}} = T_{i^\star}.
\end{equation}
This selection is performed independently for each training batch, enabling computationally advantageous while allowing fine-grained identification of the foundation model with the optimal inherent counting performance.
\subsection{Fine-tuning for multi-class cell counting}
\label{subsec1}
To further enhance the accuracy of cell quantification, we fine-tune the distilled student model, CountIHC, which has already acquired strong IHC image representations and an inherent sense of relative cell quantity through rank-aware agglomeration. Considering the clinical requirements of IHC scoring, particularly the need to quantify both positive and negative tumor cells, we design a multi-class cell counting framework driven by semantic anchors, as shown in Fig. \ref{fig:main}. It is worth noting that although our experiments primarily focus on these two categories, the framework can be naturally extended to cell counting tasks involving more categories. To reinforce category-level discrimination, we additionally introduce competitive exclusion constraints that penalize spatial overlap between predicted density distributions of different cell categories.

For the IHC image modality, we input the image into the frozen student model CountIHC to extract features. These features are then fed into a trainable vision decoder, implemented as a residual block (ResBlock) \cite{he2016deep51}, which functions as a lightweight adapter for cross-modality feature alignment. The resulting aligned feature map is denoted as $F \in \mathbb{R}^{H' \times W' \times d}$, where $H' = H // p$ and $W' = W // p$, and $d$ is the feature dimension, with $p$ denoting the fixed patch size in the ViT tokenizer \cite{dosovitskiy2020image49}.

For the text modality, we leverage the discreteness and semantic flexibility of natural language to introduce a set of discrete semantic anchors that encode both cell category and quantity information. This design draws inspiration from blockwise regression strategies in crowd counting \cite{ma2024clip34, wang2020distribution52, liu2019counting53} and is tailored to the requirements of multi-class cell counting in IHC images.

We begin by defining a category list $\mathcal{Z} = \{z_1, z_2, \ldots, z_m\}$, where $m$ denotes the total number of cell categories and $z_i$ represents the $i$-th category. For IHC images, $\mathcal{Z}$ typically includes two categories: positive tumor cell and negative tumor cell.
We then define a discrete count set $\{0, 1, \ldots, n\}$, where $n$ is a truncation threshold introduced to mitigate the influence of annotation noise and imaging artifacts. This set is converted into a collection of count bins $\mathcal{B} = \{\{0\}, \{1\}, \ldots, [n, \infty)\}$.
Each category–bin pair $(z_i, b_j) \in \mathcal{Z} \times \mathcal{B}$ is mapped to a natural language prompt using a template function $\Phi(z_i, b_j)$ defined as:
\[
\Phi(z_i, b_j) =
\begin{cases}
\text{``There is no [} z_i\text{]''} & \text{if } b_j = \{0\} \\
\text{``There is [} k \text{] [} z_i\text{]''} & \text{if } b_j = \{1\} \\
\text{``There are [} k \text{] [} z_i\text{]s''} & \text{if } b_j = \{k\},\ 2 \leq k < n \\
\text{``There are more than } n \text{ [} z_i\text{]s''} & \text{if } b_j = [n, \infty)
\end{cases}
\]
All prompts are tokenized using tokenizer of CLIP \cite{radford2021learning37} and encoded by its frozen text encoder, producing the semantic anchor tensor $A \in \mathbb{R}^{m \times (n+1) \times d}$, which incorporates both category and quantity semantics.

Similar to CLIP \cite{radford2021learning37}, we compute the cosine similarity between the encoded visual features and each semantic anchor. The resulting similarity scores are normalized via a softmax operation, yielding a probability tensor $P \in \mathbb{R}^{H' \times W' \times m \times (n+1) }$, where each sub-map $P_{i,j}(u, v)$ indicates the likelihood of count bin $b_j$ for category $z_i$ at spatial location $(u, v)$. The final multi-class density maps $D \in \mathbb{R}^{H' \times W' \times m}$ are then computed as the expectation over all count bins:
\begin{equation}
D_{i}(u, v) = \sum_{j=1}^{n} P_{i,j}(u, v) \cdot b_j
\end{equation}
where $D_{i}(u, v)$ denotes the predicted density for category $z_i$ at location $(u, v)$, and $b_j$ is the value of the $j$-th count bin.

While soft density estimation allows for continuous and overlapping responses, such overlap across semantic channels can be semantically invalid and biologically implausible. To impose inter-class exclusivity and reduce semantic interference, we propose a Spatial Exclusivity Loss (SELoss). SELoss imposes a soft mutual inhibition constraint between class-specific density maps by penalizing their high-confidence co-activation at each pixel. This regularization acts as a differentiable spatial competition mechanism, encouraging the network to resolve ambiguous activations in favor of the most semantically plausible category. Given two predicted density maps $D_1, D_2$ corresponding to different categories, the SELoss is defined as:
\[
\mathcal{L}_{se} = \frac{1}{H'W'} \sum_{u=1}^{H'} \sum_{v=1}^{W'} \sigma_\alpha(D_1(u,v) - \tau) \cdot \sigma_\alpha(D_2(u,v) - \tau)
\]
where $\sigma_\alpha(x) = \frac{1}{1 + e^{-\alpha x}}$ is a temperature-controlled sigmoid function with steepness parameter $\alpha$, and $\tau$ is a confidence threshold that filters out low-activation noise. The sigmoid-based soft filtering ensures that only confident predictions (above threshold $\tau$) are considered, while $\alpha$ controls the sharpness of exclusivity enforcement.

To further supervise both coarse-grained semantic alignment and fine-grained count accuracy, we adopt the combination of blockwise classification and distribution-matching regression losses as proposed in CLIP-EBC \cite{ma2024clip34}, where the regression component is derived from DMCount \cite{wang2020distribution54}. Moreover, to address class imbalance, we introduce a category-weighted extension of this objective, resulting in the following category-aware counting loss:
\begin{equation}
\mathcal{L}_{\text{cnt}} = \sum_{i=1}^{m} \lambda_i \left( \mathcal{L}_{\text{ce}}^{(i)} + \mathcal{L}_{\text{dm}}^{(i)} \right)
\end{equation}

\begin{equation}
\mathcal{L}_{\text{ce}}^{(i)} = - \sum_{u=1}^{H'} \sum_{v=1}^{W'} \sum_{j=1}^{n} \mathbb{1}(P_{i,j}(u,v)=1) \cdot \log \left( \hat{P}_{i,j}(u,v) \right)
\end{equation}

\begin{equation}
\mathcal{L}_{\text{dm}}^{(i)} =
\left| \left\| D_{i} \right\|_1 - \left\| \hat{D}_{i} \right\|_1 \right|
+ \mathcal{W}_2^2 \left( \frac{D_{i}}{\left\| D_{i} \right\|_1}, \frac{\hat{D}_{i}}{\left\| \hat{D}_{i} \right\|_1} \right)
+ \frac{1}{2} \left\| D_{i} \right\|_1 \left\| \frac{D_{i}}{\left\| D_{i} \right\|_1} - \frac{\hat{D}_{i}}{\left\| \hat{D}_{i} \right\|_1} \right\|_1
\end{equation}
where $\lambda_i$ is a class-specific balancing coefficient for category $i$. The cross-entropy term $\mathcal{L}_{\text{ce}}^{(i)}$ supervises the predicted probability map $\hat{P}_{i,j}(u,v)$. The indicator function $\mathbb{1}(\cdot)$ is used to focus on the pixels where the ground-truth probability is 1. The distribution matching loss $\mathcal{L}_{\text{dm}}^{(i)}$ consists of three terms: the L1 difference between the total predicted and ground-truth counts, the squared 2-Wasserstein cost $\mathcal{W}_2^2(\cdot,\cdot)$ defined as the optimal transport cost with a quadratic ground metric between the normalized density maps, and an $\ell_1$ penalty on their distributional difference. The ground-truth density maps $D_i$ and the predicted counterparts $\hat{D}_i$ are non-negative, and their normalized forms are treated as discrete probability distributions.

Combining the category-aware counting supervision and inter-class spatial exclusivity, the overall training objective is defined as:
\begin{equation}
\mathcal{L}_{\text{total}} = \mathcal{L}_{\text{cnt}} + \gamma \cdot \mathcal{L}_{\text{se}}
\end{equation}
where $\gamma$ is a hyperparameter that controls the relative weight of the exclusivity term. 



\section{Experiment}
\subsection{Evaluation protocols}
\subsubsection{Dataset}
\begin{table}[ht]
\centering
\scriptsize
\setlength{\tabcolsep}{3pt}
\renewcommand{\arraystretch}{1.2}
\caption{Summary of IHC datasets used in this study. The symbol ‘\#’ denotes the number of samples, and ‘ATC’ refers to annotated tumor cells. 
The ‘PA’ denotes whether the dataset is publicly available or not.}
\label{tab:dataset_details}
\begin{tabular}{lcccccc}
\toprule
Dataset & \# Images & \# ATC & Resolution & Biomarker(s) & Tissue & PA \\
\midrule
Ki67-Camera & 1,776 & 49,124 & 448$\times$448 & Ki-67 & Breast & $\times$ \\
Ki67-WSI & 11,409 & 832,493 & 448$\times$448 & Ki-67 & Breast & $\times$ \\
IHC-MBM & 516 & 66,196 & 448$\times$448 & 
\makecell[c]{CD11b, CD14, CD68, CD163,\\ PD-L1, TIM3, Siglec15,\\ CD73, CD86} & 
Gastric & $\times$ \\
BCData \cite{huang2020bcdata47} & 1,338 & 181,074 & 640$\times$640 & Ki-67 & Breast & $\checkmark$ \\
DeepLIIF-Data \cite{ghahremani2022deep01} & 4,484 & 98,056 & 224$\times$224 & Ki-67 & Lung, Bladder & $\checkmark$ \\
LyNSeC 1 \cite{naji2024holy55} & 1,516 & 87,316 & 224$\times$224 & CD3, Ki-67, ERG & Lymphoid & $\checkmark$ \\
\bottomrule
\end{tabular}
\end{table}
As summarized in Table~\ref{tab:dataset_details}, the datasets used in this study span 12 IHC biomarkers and 5 tissue types, including both private collections and publicly available resources. In the agglomeration stage, we utilize two Ki-67–stained private datasets, Ki67-Camera and Ki67-WSI, along with the public LyNSeC 1 \cite{naji2024holy55} dataset for unsupervised training. Subsequently, all datasets are employed to evaluate IHC counting performance, providing a comprehensive basis for assessing the proposed method across varying staining patterns and cell morphologies. Detailed dataset descriptions follow.

\noindent \textbf{\textit{IHC-Ki67.}} We collected two private Ki67-stained datasets with pixel-level annotations, comprising 618,708 negative and 262,909 positive tumor nuclei. These datasets were derived from breast tissue samples of 317 patients across Xiangya Hospital and Quanzhou Hospital. Specifically, \textbf{Ki67-Camera} was captured using a 20$\times$ microscope; \textbf{Ki67-WSI} was acquired as whole-slide images (WSIs) scanned at 20$\times$ and 40$\times$ magnification. All images were uniformly cropped into non-overlapping patches of size 448$\times$448 pixels.

\noindent \textbf{\textit{IHC-MBM.}} To validate the model's performance across a broader range of biomarker samples, we further collected 20$\times$ magnification WSIs from 31 gastric cancer patients at Zhongshan Hospital. The IHC-Multi-BioMarker (IHC-MBM) dataset includes immune markers such as myeloid markers (CD11b, CD14, CD68, CD163), immune checkpoints (PD-L1, TIM3, Siglec15), and co-stimulatory molecules (CD73, CD86). A total of 516 non-overlapping 448$\times$448 patches were cropped from the WSIs.

\noindent \textbf{\textit{BCData.}} The BCData \cite{huang2020bcdata47} consists of 1,338 Ki-67 IHC image patches (640$\times$640 pixels) extracted from 394 breast cancer WSIs, where each patch was manually annotated by pathologists with the coordinates of all tumor cell centers, which are further labeled as positive or negative based on Ki-67 staining, totally 181,074 point-level annotated tumor cells.

\noindent \textbf{\textit{DeepLIIF-Data.}} We employed the DeepLIIF dataset publicly released by Ghahremani et al.~\cite{ghahremani2022deep01}, which contains co-registered images of IHC, hematoxylin, and multiplex immunofluorescence channels derived from lung and bladder tissues. For our experiments, we extracted the IHC parts and further cropped it into 4,484 non-overlapping patches of 224$\times$224 pixels.

\noindent \textbf{\textit{LyNSeC 1.}} LyNSeC 1 \cite{naji2024holy55} includes 87,316 labeled cells across three immunomarkers (CD3, Ki-67, and ERG), annotated with nuclear contours and binary cell-level labels indicating marker-positive or marker-negative status. For our experiments, we extracted non-overlapping 224$\times$224 patches from the original images to construct the input set.

\subsubsection{Evaluation metrics}
To quantitatively evaluate the counting performance for each tumor cell category (positive and negative), we employ the mean absolute error (MAE) and root mean square error (RMSE), defined as follows:
\begin{equation}
\text{MAE} = \frac{1}{N} \sum_{i=1}^{N} \left| C_i - \hat{C}_i \right|,
\end{equation}
\begin{equation}
\text{RMSE} = \sqrt{ \frac{1}{N} \sum_{i=1}^{N} \left( C_i - \hat{C}_i \right)^2 },
\end{equation}
where $N$ is the number of images in the test set, $C_i$ is the ground-truth positive/negative tumor cell count value of image i, and $\hat{C}_i$ is the predicted count value.

Furthermore, to evaluate the model’s ability to produce clinically meaningful stratification in IHC-based analysis, we adopt the tumor proportion score (TPS), a widely used pathological metric for therapies targeting markers such as PD-L1 and HER2. It is defined as:
\begin{equation}
\text{TPS} = \frac{C_{\text{pos}}}{C_{\text{pos}} + C_{\text{neg}}},
\end{equation}
where $C_{\text{pos}}$ and $C_{\text{neg}}$ denote the number of marker-positive and marker-negative tumor cells, respectively. Based on this definition, we compute the mean absolute error of TPS predictions of each image using the same formulation as previously defined for MAE.

Lastly, we adopt the weighted mean squared error (WMSE) \cite{gao2024nwpu56} to provide a class-balanced evaluation of model performance. This metric assigns weights to the per-category MSE based on the prevalence of each cell category in the dataset, mitigating the impact of class imbalance. It is defined as:
\begin{equation}
\text{WMSE} = \frac{1}{m} \sum_{i=1}^{m} \omega_i \cdot \text{MSE}_i,
\end{equation}
\begin{equation}
\omega_i = \frac{\exp(f_i)}{\sum_{j=1}^{m} \exp(f_j)},
\end{equation}
\begin{equation}
f_i = \ln \left( \frac{\bar{C}_i}{\text{Median}(\bar{C})} \right),
\end{equation}
where $m$ is the number of cell categories, and $\text{MSE}_i$ denotes the mean squared error for category $i$. The term $\omega_i$ is a softmax-normalized weight assigned to each category, calculated from the log-ratio $f_i$, where $\bar{C}_i$ represents the total number of ground-truth cells in category $i$, and $\text{Median}(\bar{C})$ is the median count across all categories.
\subsubsection{Implementation details}
In the agglomeration stage, a total of $N = 3$ frozen teacher models are employed, specifically Virchow2 \cite{zimmermann2024virchow216}, H-Optimus-1 \cite{hoptimus117}, and CLIP-L \cite{radford2021learning37}. To align the output representations of each teacher’s image encoder with that of the text encoder, we attach an identical ResBlock \cite{he2016deep51} to each teacher. Each teacher-ResBlock pair is pretrained for 100 epochs under the same setting. This compensates for modality-specific priors in Virchow2 \cite{zimmermann2024virchow216} and H-Optimus-1 \cite{hoptimus117} for IHC, as well as the advantage of CLIP \cite{radford2021learning37} in vision-language alignment.

The student model, CountIHC, adopts the ViT-Base \cite{dosovitskiy2020image49} architecture with a patch size of $p = 14$. For each input image, a ranked patch group of size $k = 4$ is generated via center-aligned cropping, with a maximum patch size $M = 224$ and a scale ratio set to $\{\tfrac{5}{8}, \tfrac{3}{4}, \tfrac{7}{8}, 1\}$. The full framework is trained with a batch size of 128 for 200 epochs using the AdamW optimizer. The learning rate is initialized to $1 \times 10^{-3}$, decayed to a minimum of $1 \times 10^{-6}$ using a cosine annealing schedule, with linear warm-up over the first 10 epochs.

In the fine-tuning stage, the number of count bins was set to $n = 4$. The SELoss was applied with $\alpha = 10$ and $\tau = 0.3$. The category weights for count loss were set to $\lambda_1 = 0.6$ and $\lambda_2 = 0.4$ for positive and negative tumor cells, respectively, and $\gamma = 0.5$ was used to balance the Spatial Exclusivity Loss. We train our models using the Adam optimizer with a batch size of 128 and an initial learning rate of $1\times10^{-3}$, which is scheduled to decay via cosine annealing.

All experiments were conducted on 4 NVIDIA A800 GPUs. For fair comparison, all compared methods were evaluated under the same dataset protocol and trained with configurations consistent with those reported in their original papers. For detection-based methods, we fine-tuned Cellpose \cite{stringer2025cellpose3} and CellViT \cite{horst2024cellvit26}, while adopting the official checkpoints for DeepLIIF \cite{ghahremani2022deep01} (DeepLIIF\_Latest\_Model) and $\mu$SAM \cite{archit2025segment28} (vit\_h\_histopathology). As Cellpose and $\mu$SAM do not support explicit cell classification, we applied a post-hoc assignment strategy: for each predicted instance, we identified the ground-truth instance with the highest IoU (thresholded at 0.5) across all classes, and assigned the class label corresponding to the best match.
\subsection{Comparison on IHC cell counting}
\begin{table}[ht]
\caption{Quantitative results of IHC cell counting on three private datasets using five-fold cross-validation. 
Reg-based and Det-based denote regression-based and detection-based methods, respectively. Evaluation metrics include NM (MAE of negative tumor cell counts), PM (MAE of positive tumor cell counts), TM (MAE of tumor proportion score, TPS, \%), 
and WM (WMSE). The downward arrow ($\downarrow$) indicates that lower values correspond to better performance.
All values are reported as the mean across five folds. 
{\bfseries Bold} numbers denote the best results, and \underline{underlined} numbers indicate the second-best results. Best/second-best results are determined before rounding; although some rounded values appear equal, comparisons are based on the exact numbers.}
\centering
\scriptsize
\setlength{\tabcolsep}{0.8pt}
\renewcommand{\arraystretch}{1.2}
\begin{tabular}{lcllllllllllll}
\toprule
\multirow{2}{*}{Model} &
\multirow{2}{*}{Type} &
\multicolumn{4}{l}{Ki67-Camera} &
\multicolumn{4}{l}{Ki67-WSI}&
\multicolumn{4}{l}{IHC-MBM} \\
\cmidrule(r){3-6}\cmidrule(r){7-10}\cmidrule(r){11-14}
& & NM $\downarrow$ & PM $\downarrow$ &
     TM $\downarrow$ & WM $\downarrow$ & NM $\downarrow$ & PM $\downarrow$ &
     TM $\downarrow$ & WM $\downarrow$ & NM $\downarrow$ & PM $\downarrow$ &
     TM $\downarrow$ & WM $\downarrow$ \\
\midrule
CSRNet \cite{li2018csrnet} & \multirow{4}{*}{Reg-based} 
& 3.2 & {\bfseries 1.0} & 8.5 & 7.3  
& 9.9 & 2.7 & 3.8 & 52.2
& 12.6 & 6.0 & 5.0 & 85.0 \\
C-FCRN+Aux \cite{he2021deeply13} &  
& 5.1 & 1.9 & 11.9 & 15.5  
& 12.0 & 3.4 & 4.3 & 72.1 
& 21.3 & 6.6 & 6.4 & 126.4  \\
DCL \cite{zheng2024rethinking33} &  
& 3.6 & 1.4  & 12.7 & 8.6 
& 7.6 & 2.3 & 3.5 & 28.0 
& \underline{12.1} &  \underline{5.3} & {\bfseries 4.1} & \underline{74.0} \\
CLIP-EBC \cite{ma2024clip34} & 
& \underline{3.1} & \underline{1.2} & \underline{7.1} & \underline{6.1} 
& \underline{7.1} & \underline{2.1} & \underline{3.3} & \underline{24.0}  
& 12.3 & 6.5 & 5.3 & 89.2 \\
\midrule
DeepLIIF \cite{ghahremani2022deep01}  &  \multirow{4}{*}{Det-based} & 28.1  & 6.4 & 14.9 & 392.1  
& 42.6 & 6.2 & 6.0 & 431.3  
& 38.0 & 16.3 & 9.2 & 1067.8 \\
CellViT \cite{horst2024cellvit26}     &  & 7.0 & 1.9 & 9.0 & 30.7  
& 17.4 & 3.7 & 4.6 & 134.1
& 15.5 & 10.4 & 6.0 & 208.5 \\
Cellpose \cite{stringer2025cellpose3} & & 
5.4 & 2.3 & 7.2 & 21.9
& 18.2 & 4.3 & 4.5 & 143.4 
& 22.4 & 8.7 & 5.7 & 161.6 \\
$\mu$SAM \cite{archit2025segment28}   &  & 5.1 & 5.9 & 22.3 & 45.8 
& 21.0 & 14.6 & 18.7 & 376.6   
& 16.9 & 11.0 & 9.1 & 184.1  \\
\midrule
CountIHC (Ours) & Reg-based & {\bfseries 3.1} & 1.2 & {\bfseries 6.9} & {\bfseries 5.8} & {\bfseries 6.6} & {\bfseries 2.1} & {\bfseries 3.2} & {\bfseries 21.3} & {\bfseries 10.4} & {\bfseries 5.1} & \underline{4.4} & {\bfseries 58.4} \\
\bottomrule
\end{tabular}
\label{tabel_ihc_private}
\end{table}
\begin{table}[ht]
\centering
\scriptsize
\setlength{\tabcolsep}{1pt}
\renewcommand{\arraystretch}{1.2}
\caption{Quantitative results of IHC cell counting on three public datasets using five-fold cross-validation. 
Evaluation metrics include NM (MAE of negative tumor cell counts), PM (MAE of positive tumor cell counts), TM (MAE of tumor proportion score, TPS, \%), 
and WM (WMSE). The downward arrow ($\downarrow$) indicates that lower values correspond to better performance.
All values are reported as the mean across five folds. 
{\bfseries Bold} numbers denote the best results, and \underline{underlined} numbers indicate the second-best results. Best/second-best results are determined before rounding; although some rounded values appear equal, comparisons are based on the exact numbers.}
\begin{tabular}{lllllllllllll}
\toprule
\multirow{2}{*}{Model}&
\multicolumn{4}{l}{DeepLIIF-Data} &
\multicolumn{4}{l}{LyNSeC 1}&
\multicolumn{4}{l}{BCData} \\
\cmidrule(r){2-5}\cmidrule(r){6-9}\cmidrule(r){10-13}
& NM $\downarrow$ & PM $\downarrow$ &
     TM $\downarrow$ & WM $\downarrow$ & NM $\downarrow$ & PM $\downarrow$ &
     TM $\downarrow$ & WM $\downarrow$ & NM $\downarrow$ & PM $\downarrow$ &
     TM $\downarrow$ & WM $\downarrow$\\
\midrule
CSRNet \cite{li2018csrnet} 
& {\bfseries 2.2} & \underline{1.7} & 8.0 & \underline{5.5}   
& 3.2 & 2.8 & 4.8 & 14.3
& 17.5  & 8.2 & {\bfseries 3.9} & \underline{197.7} \\
C-FCRN+Aux \cite{he2021deeply13} 
& 4.7 & 3.5 & 11.9 & 15.3
& 5.2 & 4.1 & 4.4 & 26.2
& 20.7 & 9.6 & 6.8 & 308.3  \\
DCL \cite{zheng2024rethinking33} 
& \underline{2.4} & 1.7 & {\bfseries 7.1} & 5.5
& 3.5 & 3.2 & 5.2 & 17.6 
& 20.2 & \underline{8.7} & 5.1 & 277.8  \\
CLIP-EBC \cite{ma2024clip34} 
& 2.5 & 2.0 & 7.4 & 6.8
& \underline{2.9} & \underline{2.4} & \underline{3.6} & \underline{11.0}  
& \underline{16.4} & 8.8 & 5.3 & 204.0  \\
\midrule
DeepLIIF \cite{ghahremani2022deep01}  
& 7.8 & 3.1 & 11.5 & 52.3  
& 21.3 & 34.0 & 20.2 & 1206.1  
& -- & -- & -- & -- \\
CellViT \cite{horst2024cellvit26}  
& 4.6 & 4.1 & 12.4 & 30.4  
& 3.2 & 3.8 & 5.3 & 20.7  
& -- & -- & -- & -- \\
Cellpose \cite{stringer2025cellpose3}
& 8.2 & 3.8 & 9.1 & 20.5 
& 4.7 & 3.9 & {\bfseries 2.7} & 20.8 
& -- & -- & -- & -- \\
$\mu$SAM \cite{archit2025segment28}   
& 9.3 	& 6.5 & 20.9 & 37.2  
& 27.8 	& 8.6 & 53.3 & 369.6 
& -- & -- & -- & --  \\
\midrule
CountIHC (Ours)& 2.7  & {\bfseries 1.7} & \underline{7.2} & {\bfseries 5.3} 
& {\bfseries 2.7} & {\bfseries 2.3} & 3.6 & {\bfseries 10.7} 
& {\bfseries 15.5} & {\bfseries 7.4} & \underline{4.3} & {\bfseries 177.6}  \\
\bottomrule
\end{tabular}
\label{tabel_ihc_public}
\end{table}
We conduct five-fold cross-validation on six datasets spanning 5 tissues and 12 biomarker stainings to compare fine-tuned CountIHC with state-of-the-art regression-based and detection-based counting methods for IHC cell-counting accuracy. Evaluation adopts four primary metrics: NM (MAE of negative tumor cell counts), PM (MAE of positive tumor cell counts), TM (MAE of tumor proportion score, TPS), and WM (WMSE). The averaged results over the five splits are summarized in Table \ref{tabel_ihc_private} and Table \ref{tabel_ihc_public}.

Across the two IHC-Ki67 datasets we collected, CountIHC consistently demonstrates advantages under practical IHC conditions spanning different acquisition modalities. As reported in Table \ref{tabel_ihc_private}, for the class-balanced objective WM, CountIHC reduces error relative to the strongest regression baseline (CLIP-EBC \cite{ma2024clip34}) by about 4.9\% on Ki67-Camera (6.1 to 5.8) and about 7.8\% on Ki67-WSI (23.1 to 21.3). For the tumor proportion score, which is clinically important because it quantifies tumor-cell expression and guides subsequent treatment decisions, countIHC achieves lower TM than CLIP-EBC \cite{ma2024clip34} in Ki67 settings. Unlike CLIP-EBC \cite{ma2024clip34}, which relies on visual-prompt tuning of CLIP \cite{radford2021learning37} for downstream adaptation, our approach adds no extra computation overhead and performs end-to-end multi-class counting with smaller errors, indicating the effectiveness of the proposed rank-aware agglomeration strategy (RATS) together with the carefully designed multi-class fine-tuning. This advantage in balanced counting remains robust on the broader biomarker dataset (IHC-MBM) in Table \ref{tabel_ihc_private}: CountIHC lowers WM from 74.0 to 58.4 (about 21.1\%) while attaining the second-best TM of 4.4. Notably, on IHC-MBM our method achieves the lowest NM (10.4) and PM (5.1) among all compared methods. On the three public IHC datasets, as reported in Table \ref{tabel_ihc_public}, CountIHC also delivers consistent gains in WM, with reductions of approximately 3.6\% on DeepLIIF-Data (5.5 to 5.3), 2.7\% on LyNSeC-1 (11.0 to 10.7), and 10.2\% on BCData (197.7 to 177.6) over the best-performing regression baselines, while keeping TM competitive and generally lowering positive- and negative-cell MAE. In contrast, several baselines show asymmetric behavior between negative and positive tumor cells, improving one at the expense of the other, which leads to larger WM. By improving the aggregate error without favoring either class, CountIHC shows reliable performance across diverse biomarker stainings.

\begin{figure}[ht] 
  \centering                   

  \includegraphics[
    width=0.99\linewidth,       
  ]{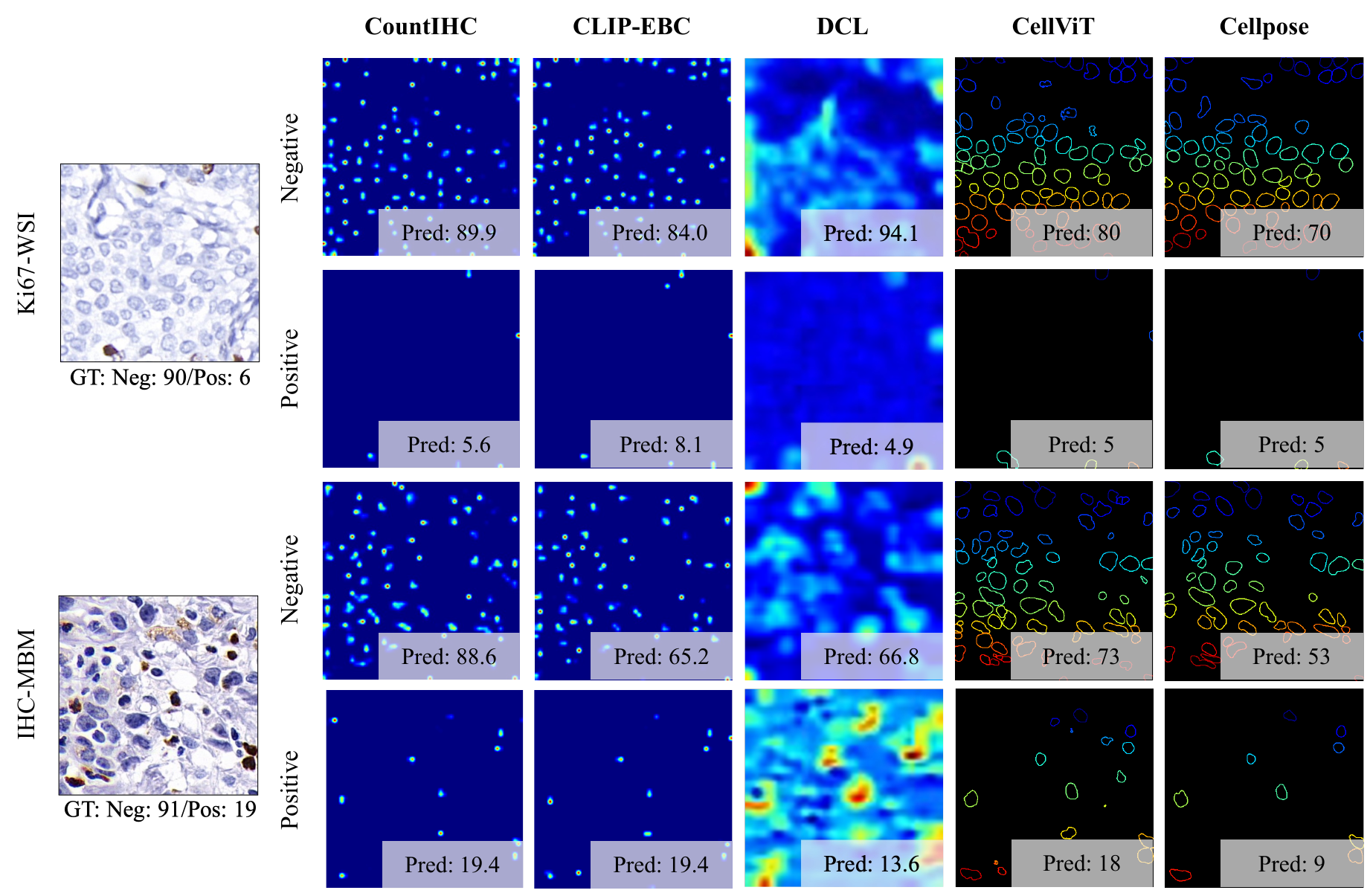}

  \caption{Qualitative comparisons of negative/positive tumor cell counting predictions on IHC images. Ground-truth counts (GT) are shown alongside the predicted counts (Pred) from different methods.}\label{fig:ihc_result}
\end{figure}

Fig. \ref{fig:ihc_result} demonstrates qualitative comparisons for the top-performing methods. For each patch, we show density maps from regression-based methods and instance contours from detection-based methods, together with the predicted negative and positive counts. CountIHC produces the closest results to the ground truth, for example, 89.9/5.6 vs. 90/6 and 88.6/19.4 vs. 91/19 (negative / positive) whereas CLIP-EBC \cite{ma2024clip34} underestimates negatives and DCL exhibits noisy responses that bias the negative-positive ratio. Detection-based approaches struggle with overlapped, small, and weakly stained cells, leading to missed counts or spurious regions. The density maps from CountIHC display well-localized peaks with clear separation between the positive and negative channels, consistent with the semantic-anchor design and the Spatial Exclusivity Loss.

\subsection{Human-machine agreement comparison}
To validate the agreement between the diagnostic outcomes of CountIHC and the clinical assessments made by pathologists, we further evaluated on two independent datasets, Ki67-Camera and Ki67-WSI, following the same source and naming convention as those used in the main experiments. Importantly, these datasets are completely separate from those used in previous experiments, thus ensuring an objective and fair comparative evaluation. Each dataset comprises cellular images from 100 patients. For diagnostic reliability, cellular patch images corresponding to each patient were organized into individual folders, with Ki67 assessments provided on the basis of 1–10 representative images per folder.

During the diagnostic process, five experienced pathologists from different hospitals participated in this study, including the First Affiliated Hospital of Xi’an Jiaotong University, Peking University People’s Hospital, the Second Affiliated Hospital of Xi’an Jiaotong University, and Quanzhou Hospital. Each pathologist independently reviewed both datasets and provided diagnostic results categorized as Ki67-, Ki67+, Ki67++, and Ki67+++. These categories correspond to increasing levels of the Ki67 labeling index (LI), which is the standard clinical measure for Ki67 expression. The LI is calculated in a manner analogous to the tumor proportion score (TPS), representing the percentage of Ki67-positive tumor cells relative to the total tumor cell population.

For the machine-based evaluation, we applied the models previously fine-tuned on Ki67-Camera and Ki67-WSI datasets to the corresponding independent test sets, Ki67-Camera and Ki67-WSI, respectively. The inference results yielded the numbers of positive and negative tumor cells, which were subsequently aggregated and used as the machine’s reference diagnostic outcomes.

\begin{figure}[ht] 
  \centering                   

  \includegraphics[
    width=0.99\linewidth,       
  ]{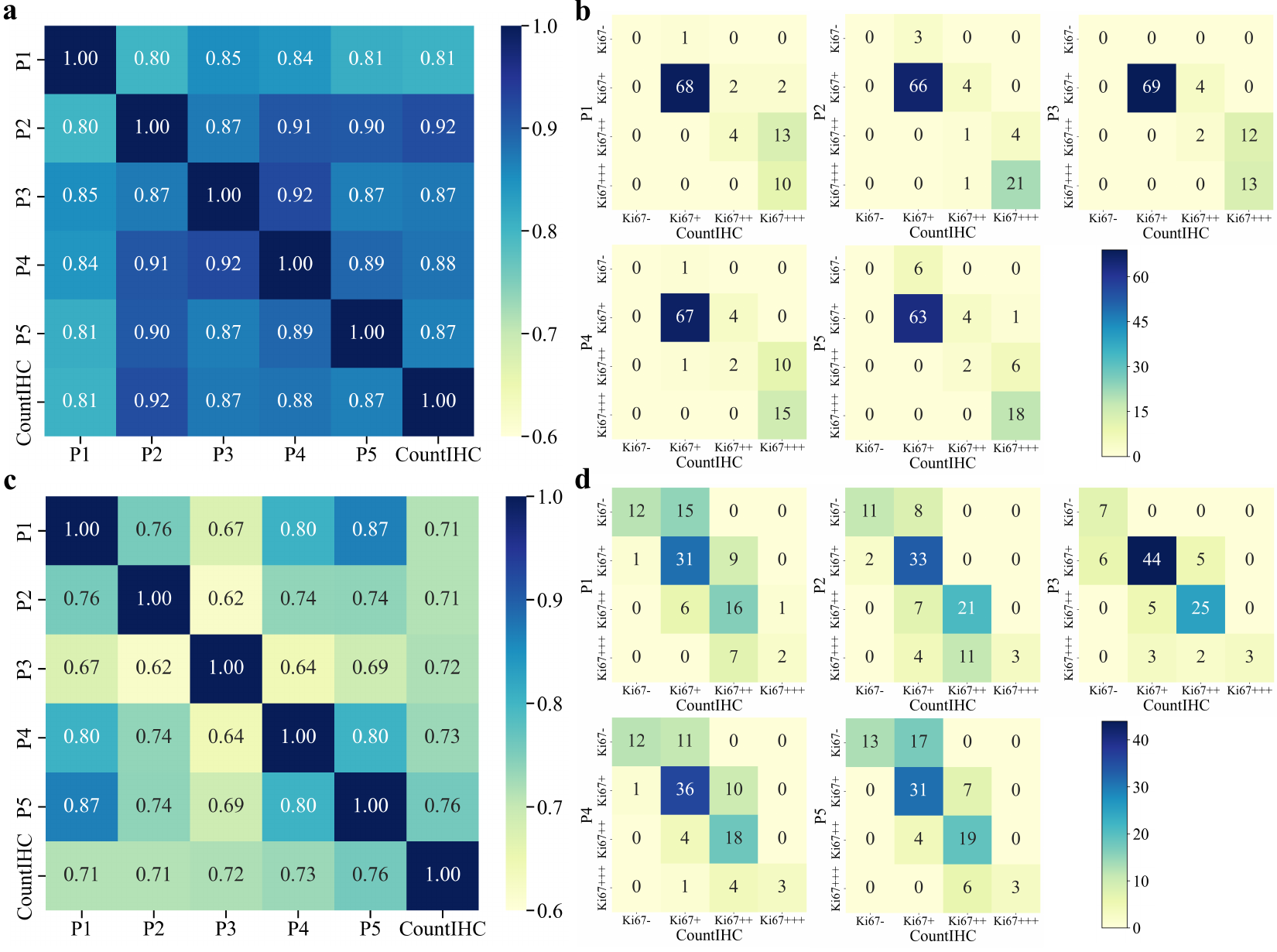}

  \caption{Agreement between the CountIHC and five experienced pathologists (P1–P5) on the Ki67-Camera and Ki67-WSI datasets. Panels \textbf{a} and \textbf{b} show the results on the Ki67-Camera dataset, with \textbf{a} presenting pairwise quadratic weighted kappa (QWK) values between the machine and pathologists and among pathologists, and \textbf{b} showing confusion matrices for grade distributions between the machine and each pathologist. Panels \textbf{c} and \textbf{d} show the corresponding analyses on the Ki67-WSI dataset.}\label{fig:cross_val}
\end{figure}

The agreement between machine and pathologists was first assessed using quadratic weighted kappa (QWK), a widely adopted metric for evaluating ordinal-scale agreement. As illustrated in Fig. \ref{fig:cross_val}, the QWK heatmaps demonstrate that on Ki67-Camera, the agreement between the machine and pathologists ranged from 0.81 to 0.92, a level closely comparable to the inter-pathologist agreement (0.80–0.92). This indicates that the model achieved human-level diagnostic reliability on camera-acquired images. On Ki67-WSI, the QWK values between the machine and pathologists were moderately lower, ranging from 0.71 to 0.76, but still within the broader range of inter-pathologist variability (0.62–0.87). These findings suggest that although the diagnostic task becomes more challenging at various magnifications, the machine remains comparable to human experts in terms of diagnostic agreement.

\begin{figure}[!ht] 
  \centering                   

  \includegraphics[
    width=0.99\linewidth,       
  ]{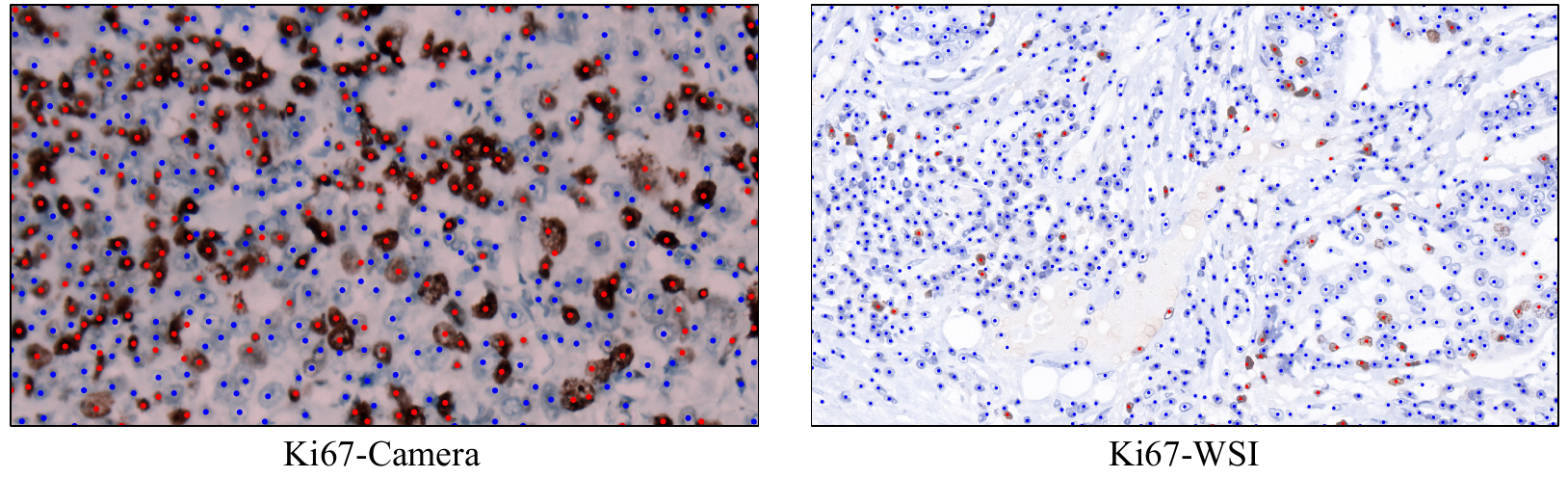}
  \caption{Visualization of approximate cell centroids derived from local maxima of the predicted class-specific density maps on Ki67-Camera and Ki67-WSI. The derived points are overlaid as markers on the tissue images, with \textcolor{blue}{\textbf{blue}} indicating negative tumor cells and \textcolor{red}{\textbf{red}} indicating positive tumor cells.}\label{fig:vis_corss}
\end{figure}

As a further assessment of diagnostic performance, confusion matrices between the machine and each pathologist are shown in Fig. \ref{fig:cross_val}. On Ki67-Camera, most cases clustered along the diagonal, indicating strong agreement across all four Ki67 expression categories. Misclassifications, when present, primarily occurred between adjacent categories (e.g., Ki67+ vs. Ki67++), which correspond to subjective boundary regions in clinical practice. On Ki67-WSI, diagonal dominance persisted but off-diagonal entries were more frequent compared with the camera dataset, particularly in intermediate categories. Notably, these discrepancies coincided with regions of reduced inter-pathologist agreement, suggesting that the machine does not introduce systematic biases but rather reflects the inherent variability among human experts. As shown in Fig. \ref{fig:vis_corss}, we further visualize local maxima of the predicted negative/positive density maps as point markers approximating cell centroids on the tissue image. This qualitative evidence substantiates the reliability of the inferred positive/negative tumor cell distributions.

These findings highlight that our proposed CountIHC achieves diagnostic performance highly consistent with that of experienced pathologists across different imaging modalities. This evidence supports the system’s potential utility as a reliable auxiliary tool in diagnostic pathology, capable of generalizing across acquisition conditions and reducing inter-observer variability in IHC scoring.
\subsection{Ablation study}
To assess whether CountIHC can inherit or even improve upon the counting ability of teacher foundation models, we use each teacher as the image encoder within our multi-class fine-tuning framework and keep the training and evaluation protocol identical to the main experiments. Without loss of generality, we evaluate on the first fold of four datasets, and we report two primary metrics that capture clinical and class-balanced accuracy: TM (MAE of the tumor proportion score) and WM (weighted mean squared error). The comparisons between the agglomerated student and Virchow2 \cite{zimmermann2024virchow216}, H-optimus-1 \cite{hoptimus117}, and CLIP-L \cite{radford2021learning37} are summarized in Table \ref{tabel_ihc_teacher}.

With parameters far fewer than any teacher, CountIHC retain or even surpass the counting performance of the best-performing teachers. On Ki67-Camera and Ki67-WSI it attains the lowest TM and WM, with WM reduced by about 9.2\% and 26.4\% relative to the strongest teacher. On the multi-biomarker IHC-MBM dataset it achieves the lowest WM, 74.2 vs. 77.6, while keeping TM competitive at 4.1, indicating better balance between negative and positive tumor cell counts under heterogeneous stainings. On BCData the student reaches the best TM of 2.8 with WM closest to the best teacher. Notably, Table \ref{tabel_ihc_teacher} also shows that the three teachers exhibit complementary strengths across the four diverse datasets. The proposed Rank-Aware Teacher Selecting (RATS) leverages this complementarity to guide agglomeration at fine granularity, producing a compact student that retains complementary teacher strengths while markedly reducing computational cost, which is advantageous for clinical deployment.
\begin{table}[ht]
\centering
\scriptsize
\setlength{\tabcolsep}{4pt}
\renewcommand{\arraystretch}{1.2}
\caption{Quantitative results of the agglomerated student model (CountIHC) and three teacher FMs on the first fold of each dataset. Evaluation metrics include TM (MAE of tumor proportion score, TPS, \%), 
and WM (WMSE). The downward arrow ($\downarrow$) indicates that lower values correspond to better performance. 
{\bfseries Bold} numbers denote the best results, and \underline{underlined} numbers indicate the second-best results. Best/second-best results are determined before rounding; although some rounded values appear equal, comparisons are based on the exact numbers.}
\begin{tabular}{
  lcllllllllll
}
\toprule
\multirow{2}{*}{Model} &
\multirow{2}{*}{\shortstack{Params $\downarrow$\\(M)}} & 
\multicolumn{2}{l}{Ki67-Camera} &
\multicolumn{2}{l}{Ki67-WSI} & \multicolumn{2}{l}{IHC-MBM} & \multicolumn{2}{l}{BCData} &\\
\cmidrule(r){3-4}\cmidrule(r){5-6}\cmidrule(r){7-8}\cmidrule(r){9-10}
& & 
     TM $\downarrow$ & WM $\downarrow$ &
     TM $\downarrow$ & WM $\downarrow$ &
      TM $\downarrow$ & WM $\downarrow$ &
      TM $\downarrow$ & WM $\downarrow$ &
          \\
\midrule
Virchow2 \cite{zimmermann2024virchow216}      & 631.2 & 6.8 & \underline{8.7}	& 4.6 & 62.7 & 4.4	& 98.6 & 3.3 & 150.0\\
H-optimus-1 \cite{hoptimus117}   & 1134.8 & \underline{6.6} & 9.6 & 3.6 & 49.6 & {\bfseries 3.2}	& 93.7 & \underline{2.9} & {\bfseries 116.6} \\
CLIP-L \cite{radford2021learning37}       & 304.0 & 9.4 & 23.2 & \underline{3.0}	& \underline{42.8} & \underline{3.73}	& \underline{77.6} & 4.4 & 207.2 \\
\midrule
CountIHC (Ours) & {\bfseries 85.7} &{\bfseries 6.1} & {\bfseries 7.9} & {\bfseries 2.5} & {\bfseries 31.5} & 4.1 & {\bfseries 74.2} & {\bfseries 2.8} & \underline{123.7}\\
\bottomrule
\end{tabular}
\label{tabel_ihc_teacher}
\end{table}

On top of CountIHC, we ablate the effect of the proposed agglomeration framework and the designed fine-tuning for multi-class cell counting on the first fold of the Ki67-WSI dataset, which involve the agglomeration strategy, i.e., Rank-Aware Teacher Selecting  (RATS), text-prompt semantic anchors, and the Spatial Exclusivity Loss (SELoss). The setting follows the main training and evaluation protocol. We report the previously defined metrics and additionally include NR and PR, the root-mean-square errors of negative and positive tumor cell counts. Results are summarized in Table \ref{tabel_alb}.

\begin{table}[ht]
\caption{Ablation study on the first fold of Ki67-WSI. 
AG denotes the agglomeration stage and FT denotes the fine-tuning stage; 
TP and RP refer to the use of text prompt and rank prompt, respectively. 
Evaluation metrics include NM (MAE of negative tumor cell counts), NR (RMSE of negative tumor cell counts), 
PM (MAE of positive tumor cell counts), PR (RMSE of positive tumor cell counts), 
TM (MAE of tumor proportion score, TPS, \%), and WM (weighted mean squared error, WMSE). 
The downward arrow ($\downarrow$) indicates that lower values correspond to better performance. 
{\bfseries Bold} numbers denote the best results, and \underline{underlined} numbers indicate the second-best results. 
w/ and w/o denote “with” and “without,” respectively.
Best/second-best results are determined before rounding; although some rounded values appear equal, 
comparisons are based on the exact numbers.}
\centering
\scriptsize
\setlength{\tabcolsep}{4pt}
\renewcommand{\arraystretch}{1.2}

\begin{tabular}{
  c|cllllll
}
\toprule
\multicolumn{2}{c}{\multirow{2}{*}{Setting}} 
 & \multicolumn{6}{c}{Ki67-WSI} \\
\cmidrule(lr){3-8}
\multicolumn{2}{c}{} & NM $\downarrow$ &
     NR $\downarrow$ & PM $\downarrow$  & PR $\downarrow$ & 
     TM $\downarrow$ & WM $\downarrow$ \\
\midrule
\multirow{2}{*}{\shortstack{Agglomeration\\strategy}}
& Equal learning \cite{ma2024towards19, ranzinger2024radio20, heinrich2025radiov222} & 8.0 	& 12.0 	& 4.3 	& 5.7 	& 4.3 	& 48.1  \\
& tdrop \cite{sariyildiz2024unic21, sariyildiz2025dune23} & 7.9 	& 12.3 	& 3.4 	& 5.1 	& 3.6 	& 36.3  \\
\midrule
\multirow{2}{*}{\shortstack{Prompt}}
& AG-TP\&FT-RP & 7.5 	& \underline{11.8} 	& 2.9 	& 4.6 	& 2.7 	& 34.1  \\
& AG-RP\&FT-RP & 7.8 	& 11.9 	& 2.9 	& 4.4 	& 2.7 	& \underline{34.1}  \\
\midrule
Loss & w/o SELoss & \underline{7.4} 	& 11.9 	& \underline{2.7} 	& \underline{4.3} 	& \underline{2.5} 	& 35.4  \\
\midrule
Ours & \parbox[c]{2.5cm}{\centering RATS\\AG-TP+FT-TP\\w/ SELoss} & {\bfseries 7.4} & {\bfseries 11.7} & {\bfseries 2.6}	& {\bfseries 4.1} & {\bfseries 2.5}	& {\bfseries 31.5} \\
\bottomrule
\end{tabular}
\label{tabel_alb}
\end{table}
For the agglomeration stage, we compare RATS with two widely known agglomeration strategies, equal learning \cite{ma2024towards19, ranzinger2024radio20, heinrich2025radiov222} and teacher dropping regularization (tdrop) \cite{sariyildiz2024unic21, sariyildiz2025dune23}, using the same set of teachers and fine-tuning settings. Compared with these strategies, RATS reduces the WM from 48.1 and 36.3 to 31.5 and lowers the TM from 4.3 and 3.6 to 2.5, accompanied by consistent reductions in NM, PM, NR, and PR. We attribute this to a key limitation of these strategies: the absence of an explicit assessment of teacher capacity prevents them from adaptively allocating supervision according to task-specific performance. In contrast, RATS performs task-driven agglomeration by dynamically identifying the optimal teacher based on its inherent counting ability, thereby assigning more appropriate supervision and achieving improved performance.

\begin{figure}[ht] 
  \centering                   

  \includegraphics[
    width=0.99\linewidth,       
  ]{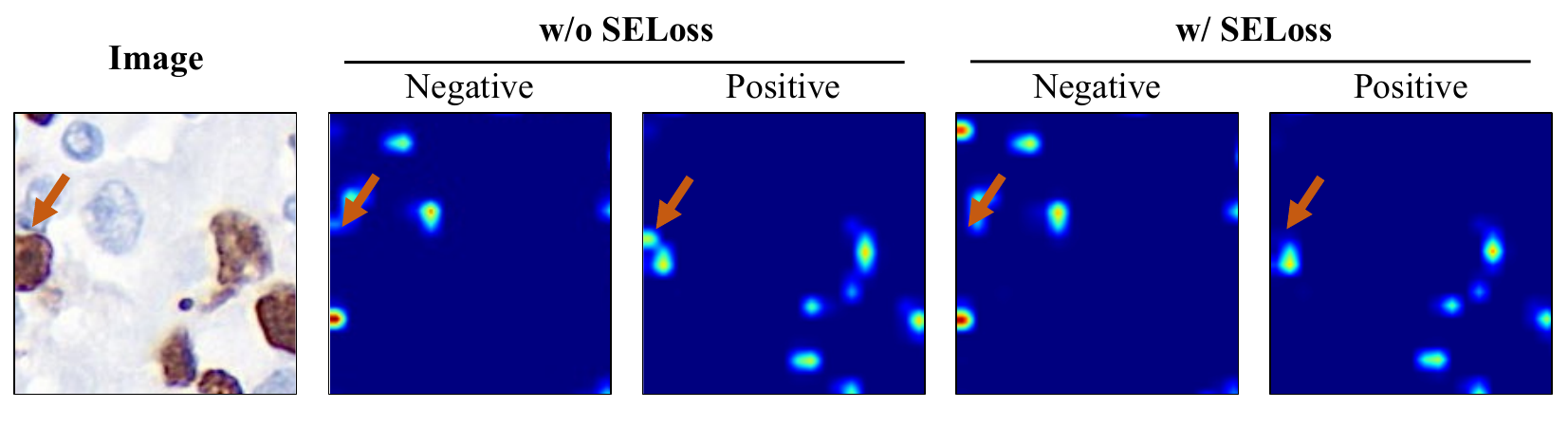}

  \caption{Visualization of SELoss effects on class-specific density maps. Without SELoss, the negative and positive channels spuriously co-activate at the same locations. With SELoss, this cross-class co-activation is suppressed and the positive responses at the arrowed locations are sharper and better separated.}\label{fig:alb}
\end{figure}

For semantic anchors, we compare text prompts and rank prompts at both the agglomeration (AG) and fine-tuning (FT) stages. Building on \cite{li2022ordinalclip}, which emphasizes a rank-prompt formulation, we adapt the method by introducing two class-specific rank prompters (one for negative and one for positive tumor cells), while keeping all other settings identical to the optimal configuration reported therein. As shown in Table \ref{tabel_alb}, replacing text prompts with rank prompts at either AG or FT yields inferior results. Using text prompts at both stages provides a clear, class-aware alignment signal without introducing additional modules and achieves the strongest performance, lowering the WM from 34.1 to 31.5 and improving the TM from 2.7 to 2.5. The introduction of SELoss further brings a consistent improvement, reducing the WM from 35.4 to 31.5, with concurrent decreases in NR and PR. By penalizing high-confidence co-activation between class-specific density maps at each pixel, SELoss mitigates cross-class confusion and improves both TPS accuracy and the overall balanced error. As shown in Fig. \ref{fig:alb}, training without SELoss leads to erroneous co-activation of the negative and positive channels at the same locations, whereas SELoss suppresses this cross-class co-activation and yields sharper, better-separated positive responses at the same locations highlighted with arrows, confirming the intended effect of the loss.
\subsection{Applicability to H\textsl{\&}E-stained images}
To further validate the scalability of our proposed agglomeration framework and fine-tuning strategy, we extended the experiments to H\&E-stained images, another widely used modality in clinical pathology. During the agglomeration stage, we adopted seven publicly available datasets (CoNSeP \cite{graham2019hover32}, CPM17 \cite{vu2019methods58}, LyNSec 2\&3 \cite{naji2024holy55}, MoNuSeg \cite{kumar2019multi59}, PanNuke \cite{gamper2019pannuke60}, NuInsSeg \cite{mahbod2024nuinsseg61}, and PUMA \cite{schuiveling2025novel62}) to conduct rank-aware agglomeration of foundation models, in which the same three foundation models (Virchow2 \cite{zimmermann2024virchow216}, H-optimus-1 \cite{hoptimus117} and CLIP-L \cite{radford2021learning37}) were distilled into a single student model without relying on manual annotations. The data preprocessing pipeline, methodological settings, and training procedures were kept consistent with those described for IHC-stained images. At the fine-tuning stage, we employed MoNuSeg \cite{kumar2019multi59} (cropped into non-overlapping 224$\times$224 patches) and TNBC \cite{naylor2018segmentation57} (cropped into non-overlapping 448$\times$448 patches) to adapt the student model, and subsequently evaluated its cell counting performance on H\&E-stained images.
\begin{table}[ht]
\caption{Quantitative results of cell counting on H\&E-stained datasets (MoNuSeg and TNBC). Evaluation metrics include mean absolute error (MAE) and root mean squared error (RMSE). The downward arrow ($\downarrow$) indicates that lower values correspond to better performance. {\bfseries Bold} numbers denote the best results, and \underline{underlined} numbers indicate the second-best results.}
\centering
\scriptsize
\setlength{\tabcolsep}{4pt}
\renewcommand{\arraystretch}{1.2}

\begin{tabular}{
  lllll
}
\toprule
\multirow{2}{*}{Model} &
\multicolumn{2}{l}{MoNuSeg} &
\multicolumn{2}{l}{TNBC} \\
\cmidrule(r){2-3}\cmidrule(r){4-5}
& \multicolumn{1}{l}{MAE} $\downarrow$ &
  \multicolumn{1}{l}{RMSE} $\downarrow$ &
  \multicolumn{1}{l}{MAE} $\downarrow$ &
  \multicolumn{1}{l}{RMSE} $\downarrow$ \\
\midrule
Virchow2 \cite{zimmermann2024virchow216}      & 3.1 & 4.0 & 7.8 & 8.9 \\
H-optimus-1 \cite{hoptimus117}   & 3.1 & 3.9 & \underline{5.5} & {\bfseries 6.7} \\
CLIP-L \cite{radford2021learning37}       & 3.5 & 4.7 & 6.6 & 8.2 \\
\midrule
CSRNet \cite{li2018csrnet}        & 3.2 & 4.4 & 7.0 & 8.1 \\
C-FCRN+Aux \cite{he2021deeply13}    & 5.1 & 6.5 & 12.1 & 14.2 \\
DCL \cite{zheng2024rethinking33}           & 4.0 & 5.2 & 11.5 & 12.3 \\
CLIP-EBC \cite{ma2024clip34}     & 3.3 & 4.4 & 15.2 & 17.1 \\
CellViT \cite{horst2024cellvit26}      & 3.8 & 5.1 & 16.1 & 19.6 \\
Cellpose \cite{stringer2025cellpose3}    & \underline{3.0} & \underline{3.8} & 11.7 & 13.4 \\
\midrule
CountIHC (Ours) & {\bfseries 2.7} & {\bfseries 3.4} & {\bfseries 5.0} & \underline{7.8} \\
\bottomrule
\end{tabular}
\label{tabel_he}
\end{table}

Quantitative results are presented in Table \ref{tabel_he}. On MoNuSeg, CountIHC achieved the lowest MAE of 2.7 and RMSE of 3.4, improving over the compared method Cellpose (MAE 3.0 / RMSE 3.8) by 10.0\% and 10.5\%, respectively. On TNBC, CountIHC reached an MAE of 5.0 and RMSE of 7.8, which are competitive with H-optimus-1 \cite{hoptimus117} (MAE 5.5 / RMSE 6.7), the best-performing teacher FM, while requiring significantly less computation overhead. These results echo the findings on IHC-stained images, demonstrating that our approach consistently retains or even surpasses the counting performance of the best-performing teachers across diverse histopathological modalities.
\begin{figure}[!h] 
  \centering                   

  \includegraphics[
    width=0.99\linewidth,       
  ]{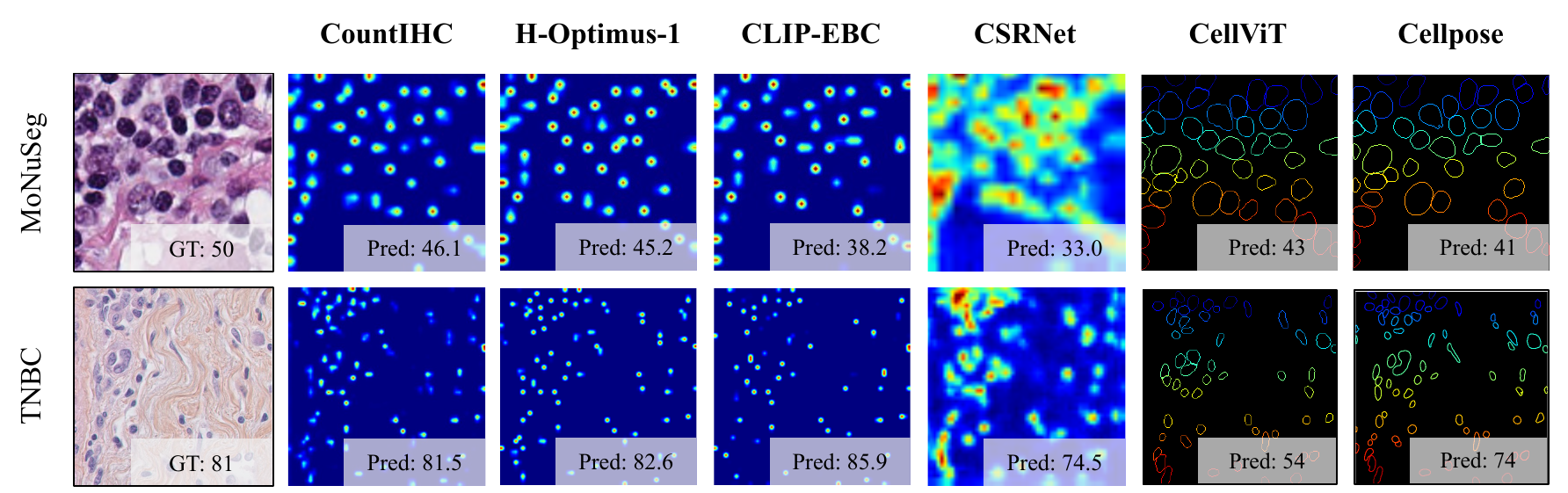}

  \caption{Qualitative comparisons of cell counting predictions on H\&E-stained images from MoNuSeg (top) and TNBC (bottom). Ground-truth counts (GT) are shown alongside the predicted counts (Pred) from different methods.
}\label{fig:he_result}
\end{figure}

The qualitative comparisons in Fig. \ref{fig:he_result} further support these findings. CountIHC yields estimates that remain closely aligned with the ground-truth values (e.g., Pred: 46.1 vs. GT: 50 on MoNuSeg, Pred: 81.5 vs. GT: 81 on TNBC). In contrast, detection-based methods (e.g., CellViT, Cellpose) tend to under-segment or miss densely packed nuclei, while regression-based models (e.g., DCL, CSRNet) often generate overly smoothed or noisy predictions, leading to significant under- or over-estimation.

Both quantitative and qualitative analyses verify that the proposed agglomeration framework and fine-tuning strategy extend robustly from IHC- to H\&E-stained images, underscoring their scalability for reliable cell counting across modalities.

\section{Discussion}
\begin{figure}[ht] 
  \centering                   

  \includegraphics[
    width=0.85\linewidth,       
  ]{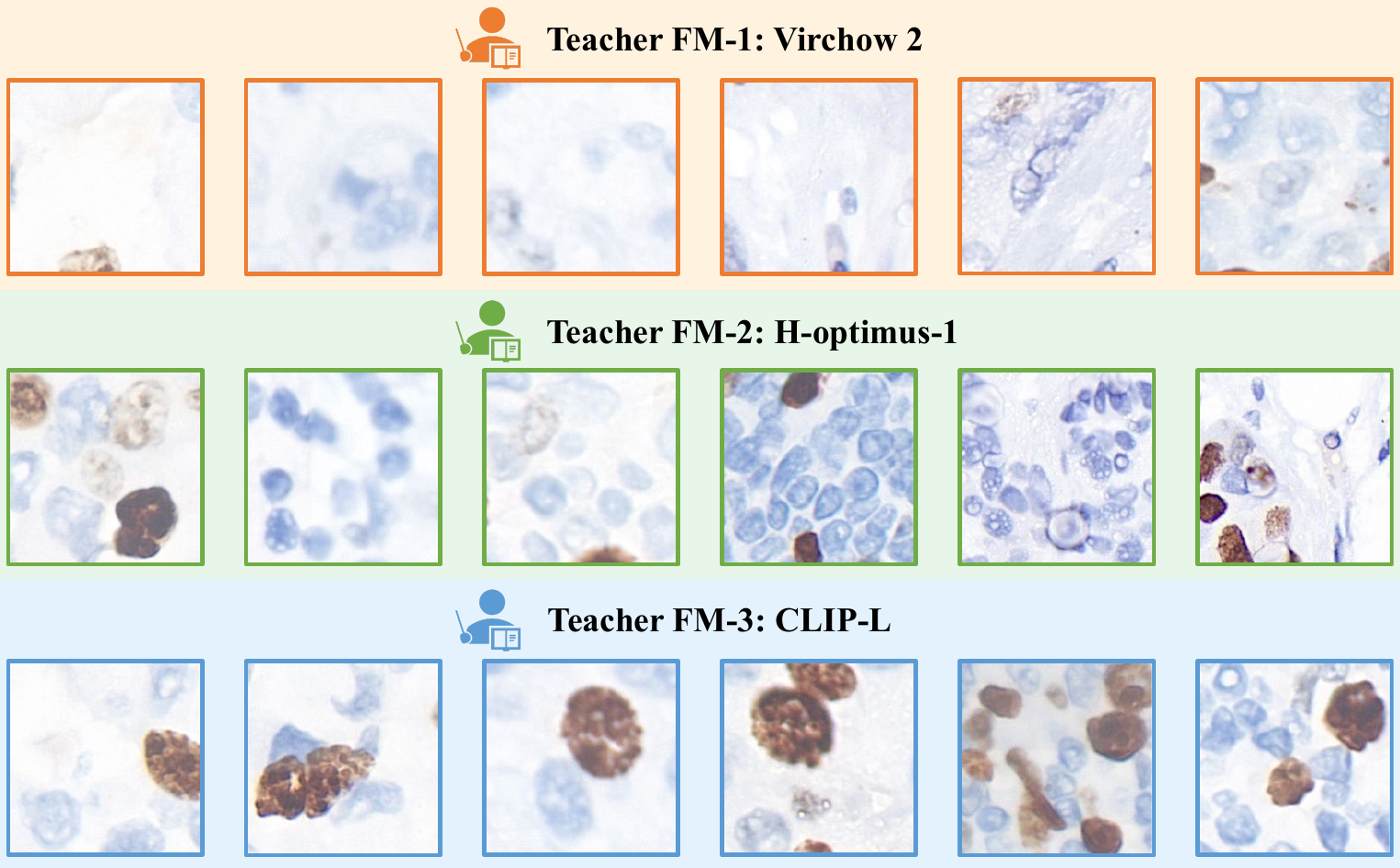}

  \caption{Visualization of sample-level teacher FM assignments by RATS.}\label{fig:discussion}
\end{figure}
Our work presents a rank-aware agglomeration framework that distills
complementary strengths from multiple foundation models into a compact
student, CountIHC. To further illustrate the complementary counting capacities of different teachers in IHC, Fig. \ref{fig:discussion} visualizes sample-level teacher assignments produced by the proposed Rank-Aware Teacher Selecting (RATS) strategy. In practice, we observe that H-Optimus-1 \cite{hoptimus117} is the most frequently selected teacher, particularly for samples with crowded nuclei, pronounced morphological heterogeneity, or weak and uneven staining. Its larger model size and broader pretraining data scale provide more robust supervision, consistent with its role as the overall best-performing teacher reported in Table \ref{tabel_alb}. Virchow2 \cite{zimmermann2024virchow216} is more often selected in low-contrast or sparsely cellular regions, where conservative responses help suppress spurious positives and stabilize the negative-cell channel. CLIP-L \cite{radford2021learning37} is preferentially assigned to patches containing prototypical, sharply stained positive tumor nuclei with clear boundaries, where precise localization is critical. The observed teacher-specific assignments arise from the RATS strategy, which employs an unsupervised global-to-local patch ranking task aligned with cell counting. It distills knowledge from the teacher exhibiting higher rank consistency within each patch group, as this indicates stronger inherent counting capability. Collectively, the results in Fig. \ref{fig:discussion} confirm that RATS adaptively leverages the complementary strengths of different teachers, allowing CountIHC to remain compact while retaining or even surpassing the performance of the best single teacher.

Combined with semantic anchors that encode both cell quantity and category information through structured text prompts, and a Spatial Exclusivity Loss (SELoss) that enforces inter-class spatial separability, the framework aligns visual features with discrete semantic representations, enabling accurate and category-aware density estimation. Across 6 multi-center IHC datasets covering 12 biomarker stainings and 5 tissue types, five-fold cross-validation against state-of-the-art detection- and regression-based methods yields consistently superior performance on most metrics. A human–machine agreement study further shows a high degree of agreement with pathologists’ assessments, underscoring reliability and potential clinical value. Ablations verify that both the proposed agglomeration and the fine-tuning strategy contribute materially to the gains. Finally, transferring the method to H\&E-stained data indicates encouraging modality robustness and scalability.

Looking ahead, we plan to scale the data used during agglomeration and extend evaluation beyond patch-level counting to slide- and region-level tasks that are common in IHC practice and benchmarks. These include biomarker scoring such as H-score estimation or PD-L1 CPS, immune contexture analysis such as compartment-specific TIL density, cell phenotyping or subtype classification, and fine-grained instance-level cell segmentation. These studies will further characterize the breadth of capabilities inherited through rank-aware multi-teacher distillation and assess downstream clinical utility.

\section{Conclusion}
In this work, we present CountIHC, a model derived from rank-aware agglomeration of multiple foundation models, achieving comparable or superior performance to the best teacher FM in IHC cell counting, with substantially reduced computation overhead. Building on the complementary representations of foundation models, CountIHC effectively handles heterogeneous staining and morphological patterns in IHC. Specifically, through the Rank-Aware Teacher Selecting (RATS) strategy, CountIHC dynamically selects the optimal teacher sample-wise via an unsupervised patch ranking task, facilitating adaptive and task-driven knowledge learning. Additionally, CountIHC undergoes a fine-tuning stage that reformulates counting as vision–language alignment. In this stage, discrete semantic anchors, constructed through structured text prompts, encode both cell category and quantity information. Their alignment with image features regresses class-specific density maps and improves counting accuracy in cell overlapping regions. Extensive experiments across diverse IHC datasets demonstrate that CountIHC outperforms state-of-the-art methods and achieves high agreement with pathologists’ assessments, highlighting its potential clinical value. Our method also demonstrates its scalability when applied to H\&E-stained data.






\section*{Declaration of Generative AI and AI-assisted technologies in the manuscript preparation process}
During the preparation of this work, the authors used ChatGPT for language polishing and grammar refinement only. After using this tool, the authors reviewed and edited the content as needed and take full responsibility for the content of the published article.
\bibliographystyle{elsarticle-num} 
\bibliography{references}






\end{document}